\documentclass[runningheads]{llncs}
\usepackage[T1]{fontenc}
\usepackage{graphicx}
\usepackage{array, multirow, booktabs, ragged2e, tabularx, amssymb}
\usepackage{subfigure, caption}
\newcommand{\footnoteref}[1]{\textsuperscript{\ref{#1}}}

\begin{document}
\title{A Performance Increment Strategy for Semantic Segmentation of Low-Resolution Images from Damaged Roads}

%
%

\author{Rafael S. Toledo\inst{1}\thanks{Corresponding Author}, Cristiano S. Oliveira \inst{1}, Vitor H. T. Oliveira \inst{1}, Eric A. Antonelo\inst{1}, Aldo von Wangenheim\inst{1}}

%
\authorrunning{F. Author et al.}
%


\institute{Federal University of Santa Catarina, Florianópolis, Santa Catarina, Brazil}

\maketitle              
\begin{abstract}
Autonomous driving needs good roads, but 85\% of Brazilian roads have damages that deep learning models may not regard as most semantic segmentation datasets for autonomous driving are high-resolution images of well-maintained urban roads. A representative dataset for emerging countries consists of low-resolution images of poorly maintained roads and includes labels of damage classes; in this scenario, three challenges arise: objects with few pixels, objects with undefined shapes, and highly underrepresented classes. To tackle these challenges, this work proposes the Performance Increment Strategy for Semantic Segmentation (PISSS) as a methodology of 14 training experiments to boost performance. With PISSS, we reached state-of-the-art results of 79.8 and 68.8 mIoU on the Road Traversing Knowledge (RTK) and Technik Autonomer Systeme 500 (TAS500) test sets, respectively. Furthermore, we also offer an analysis of DeepLabV3+ pitfalls for small object segmentation. \footnote{Code available on \url{https://github.com/tldrafael/pisss}.}

\end{abstract}
\section{Introduction}

Autonomous driving research is mainly based on developed countries with well-maintained infrastructure represented by many European urban streets datasets of high-res images, e.g., Cityscapes \cite{cordts2016cityscapes}, CamVid \cite{brostow2009semantic}, and KITTI \cite{geiger2013vision}. In Brazil, 85\% of the roads suffer from fatigue, cracks, holes, patches, and wavy surfaces \cite{CNT2021}. This poor condition demands adjustments in the perception of autonomous driving. Additionally, computational constraints in emerging countries may limit the input size for the deep learning models, constraining the usage of high-res images and forcing adapted solutions for low-res images.

\begin{figure}[hbt!]
\centering
\includegraphics[width=\textwidth]{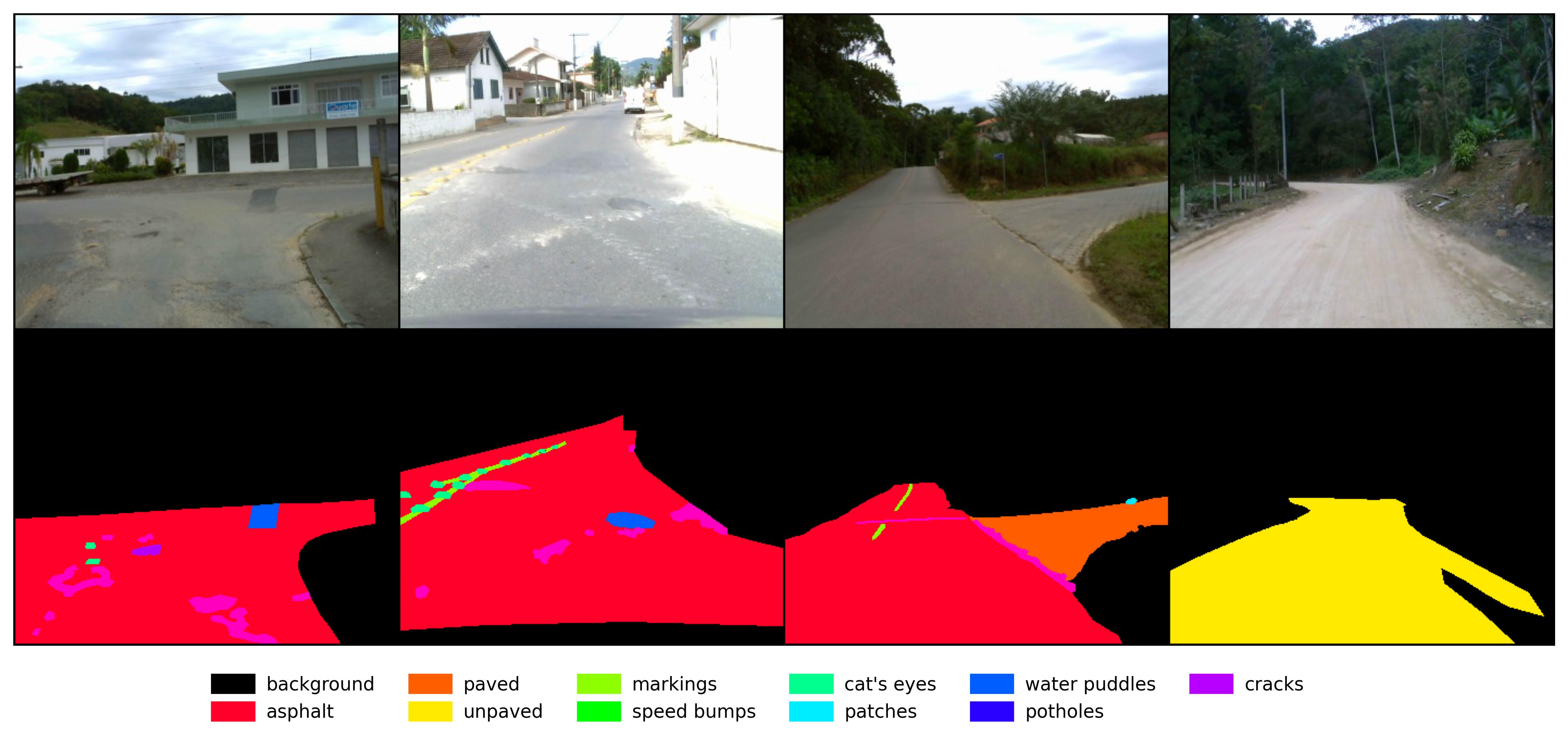}
\caption{Examples of emerging countries' roads in the RTK dataset.} \label{fig:ch3_RTK_samples_1}
\end{figure}

Some datasets such as  \cite{rateke2019road,shinzato2016carina,varma2019idd} represented emerging countries' roads. Among them, especially the RTK dataset \cite{rateke2019road} captures the Brazilian countryside road featuring distinct maintenance conditions and surfaces. RTK consists of 701 annotated images of resolution 352x288 with 12 classes. The classes include surfaces (asphalt, paved, and unpaved), signs (markings, cat's eyes, speed bumps, and storm drains), and damages (patches, water-puddles, potholes, and cracks). See Figure \ref{fig:ch3_RTK_samples_1}. 

This work raised three challenges for training a deep learning model when working with low-res images: objects with few pixels, objects with undefined shapes, and highly underrepresented classes. Low-res images have small objects not only relative to other objects' sizes but in pixels' quantity; for example, 70\% of the cat's eyes blobs have edges with equal or less than 5 pixels, and 15\% of the road markings blobs even have edges of a unique pixel. These tiny objects can easily vanish at the beginning of the forward pass given the stride of convolutional and pooling layers. The vanishing objects became a problem for DeepLabV3+, as will be seen in Sec. \ref{dv3_patt}.

Other issues are undefined shape objects or multiscale elements presence in the image, e.g., road surfaces are broad and have a well-defined shape, whereas patches do not have a defined shape and size. This problem complexity increases with intraclass shape variations, like holes and cracks in the same image with multiple formats. Lastly, small damages and signs become very underrepresented, e.g., background and road surfaces occupy 98.5\% of pixels, while cat's eye and storm-drain are just 0.02\%. The imbalanced scenario raises the risk of overlooking small-sized classes.

\subsection{Contributions}

To meet the challenges presented above, we designed the Performance Increment Strategy for Semantic Segmentation (PISSS) that consists of a series of good training practices found in the state-of-the-art (SOTA) works that tackle semantic segmentation challenges of imbalanced datasets, small objects, and multiscale segmentation.  
With PISSS, we raised the RTK benchmark to 79.8 mIoU and the TAS500 to 68.8 mIoU, the best published results so far. Furthermore, we also propose removing the ResNet's max-pooling (MP) layer to preserve small objects segmentation.

\section{Background}

In this section, we introduce the main topics covered in the PISSS strategy, they are: training procedures, small objects segmentation, and multiscale segmentation.

\subsection{Guidelines and Training Procedures}
A significant part of deep learning success comes from adopting better training procedures. However, they are not usually the main research focus, and their details may be hidden in the implementation code. The Bag of Tricks for Image Classification  \cite{he2019bag} is a fundamental work for training recipes with convolutional neural networks, and it pushed ResNet-50's ImageNet top-1 validation accuracy from 75.3\% to 79.29\%. \cite{wightman2021resnet,bello2021revisiting} also present common procedures like warm-up learning rate (LR), cosine LR decay, weight decay, label smoothing, stochastic depth, dropout, mixup, cutmix, and random resized cropping.

Furthermore, standard practices in semantic segmentation SOTA works \cite{chen2018encoder,zhao2017pyramid,yuan2019segmentation,kirillov2020pointrend,yu2021bisenet} are cropping, resizing, and flipping as data augmentation; Stochastic Gradient Descent (SGD) with a polynomial LR decay, momentum of 0.9, and weight decay of 5e-4 as optimizer; and multiscale ensemble predictions for testing. The SGD preference over Adam is explained by its better generalization results \cite{keskar2017improving,zhou2020towards}.

Important takeaways are pointed out by \cite{wightman2021resnet} that there is no training procedure ideal for all models, and by \cite{bello2021revisiting} that training methods are more task-driven than architectures, and, hence, improvements from training methods do not necessarily generalize as well as architectural ones.

\subsection{Small Objects Segmentation}

Small object segmentation is a challenge for low-res datasets. \cite{hu2017finding,hamaguchi2018effective} emphasized the importance of context for small object detection, underlining that even humans cannot recognize a small building in a satellite image without the context of roads, cars, or other buildings. Hence, neurons with a large receptive field (RF) are essential for the task. A simpler option was found by \cite{shen2017dsod} that noticed that the double sequence of stride 2 on ResNet loses sensible feature information of small objects; they proposed replacing the ResNet 7x7 convolution layer with stride 2 to a series of three 3x3 with stride only at the end. This same tactic was followed later by \cite{zhou2018scale,li2019road}. 

\subsection{Multiscale Segmentation and DeepLabV3+}

Some approaches for handling the multiscale segmentation challenge are to extract multiscale features in a layer level like Res2Net \cite{gao2019res2net} and to add an attention mechanism that smartly combines predictions from different feature map scales to avoid scale pitfalls \cite{tao2020hierarchical,chen2016attention}. \cite{tao2020hierarchical} noted that feature maps of large scales predict better fine details, such as edges of objects or thin structures, whereas feature maps of small scales predict better large structures that demand global context.

DeepLabV3+ \cite{chen2017deeplab} proposed the Atrous Spatial Pyramid Pooling (ASPP) module in charge of capturing multiscale features by simultaneously applying various dilated convolutions, hence, combining multiple receptive fields. ASPP balances the trade-off between accurate localization (small receptive field) and context assimilation (large receptive field).

\section{PISSS - Performance Increment Strategy for Semantic Segmentation} \label{sec_solution}

PISSS is a methodology that consists of an additive series of ablation experiments. Each experiment checks the best performance among a set of hypotheses; e.g., which augmentation operation works better for the RTK dataset? Geometry operation? Color operations? Both together? Each option is called a hypothesis. The ablation experiments answer the questions of the best hypothesis for part of the training setup. The next ablation experiment is built on top of the best setup until that moment.

\begin{figure}[hbt!]
\centering
\includegraphics[width=.9\textwidth]{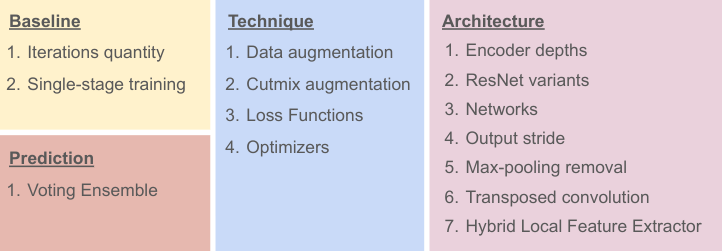}
\vspace*{-3mm}
\caption{PISSS Diagram.} \label{fig:PISSS_diagrama}
\end{figure}

In total, the PISSS applied on the RTK dataset sums 14 ablation experiments organized into four categories: Baseline (B), Prediction (P), Technique (T), and Architecture (A). Baseline checks choices of the RTK authors training in \cite{rateke2021road}, Prediction checks the usage of prediction ensemble, Technique checks training setup specificities, and Architecture checks changes in the neural network structure. Figure \ref{fig:PISSS_diagrama} shows the experiments tried in each category.

The sequence of experiments for PISSS starts on the baseline work, and we also adopted a different evaluation strategy as discussed next.

\subsection{Baseline}

We adopted the RTK authors' solution \cite{rateke2021road} as the starting point for the sequence of experiments. We call it \textbf{baseline}. It consists of a U-Net with ResNet-34, Adam optimizer with LR of 1e-4, batch size of 8, data augmentation with perspective distortion and horizontal flipping, named as \textit{GeomRTK}, and a two-stage training regime that first runs 100 epochs with cross-entropy (CE) and another 100 later with weighted cross-entropy (WCE), summing 14k iterations at total.

\subsection{Evaluation Methodology}

\begin{figure}[hbt!]
\centering
\includegraphics[width=.4\textwidth]{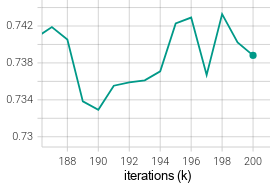}
\vspace*{-3mm}
\caption{mIoU oscillation after loss convergence.} \label{fig:ch4_methodology_1}
\end{figure}

When monitoring our training experiments, we noticed slight mIoU variations after the loss convergence. In Fig. \ref{fig:ch4_methodology_1}, it is seen an oscillation around [0.733,0.743], which represents 1\% of the mIoU scale of [0,1]; this variation is enough to lead to an incorrect conclusion when comparing the hypotheses. We adopted a workaround to reduce this noisy variation by averaging the last ten results steps of the validation set.

\section{Applying PISSS on RTK}

In this section, we present the PISSS application over the RTK\footnote{\url{https://data.mendeley.com/datasets/hssswvmjwf/1}}. We split it into 5 main parts that most leverage performance, following the chronological order: 1) surpassing the baseline, 2) tuning DeepLabV3+, 3) fancy approaches and cutmix, 4) loss functions, and 5) prediction ensemble. 

\subsection{Part 1 - Surpassing the Baseline}

In the first part, we covered the five first experiments: Iterations (B), Single Stage Training (B), Data Augmentation (T), Encoder Depths (A), and Resnet Variants (A). See Table \ref{tab:abl_rkt_1}; the order follows from top to bottom. Every subsequent experiment is built over the best hypothesis until that moment unless it is otherwise said, and the best hypothesis until that time has its mIoU value in bold.  

\begin{table}[hbt!]
\centering
\caption{Part 1 of PISSS.}
\vspace{.15cm}
\begin{tabular}{l|l|l|l|l|l}
\hline
Exp. & Enc. & DataAug & Losses & Iters & mIoU \\
\hline
B: Iterations & ResNet-34 & GeomRTK & CE+WCE &  14k & 54.7 \\
B: Iterations & ResNet-34 & GeomRTK & CE+WCE & 100k & 69.3 \\
B: Iterations & ResNet-34 & GeomRTK & CE+WCE & 200k & \textbf{73.9} \\
\hline
B: Single-stage & ResNet-34 & GeomRTK & WCE & 200k & 73.2 \\
B: Single-stage & ResNet-34 & GeomRTK & CE & 200k & \textbf{73.9} \\
\hline
T: DataAug & ResNet-34 & None & CE & 200k & 66.8 \\
T: DataAug & ResNet-34 & Cutmix80\footnotemark & CE & 200k & 71.0 \\
T: DataAug & ResNet-34 & Cutmix50\footnoteref{fn:cutmix_1} & CE & 200k & 71.0 \\
T: DataAug & ResNet-34 & Resizing+Crop. & CE & 200k & 73.5 \\
T: DataAug & ResNet-34 & Crop+Color & CE & 200k & 75.1 \\
T: DataAug & ResNet-34 & Crop & CE & 200k & \textbf{75.4} \\
\hline
A: Encoder Depth & ResNet-50 & Crop & CE & 200k & 75.9 \\
A: Encoder Depth & ResNet-101 & Crop & CE & 200k & \textbf{76.5} \\
\hline
A: ResNet Variants & Res2Net-101 & Crop & CE & 200k & 74.6 \\
A: ResNet Variants & ResNeXt-101 & Crop & CE & 200k & 75.4 \\
A: ResNet Variants & ResNeSt-101 & Crop & CE & 200k & 75.9 \\
\hline
\end{tabular}
\label{tab:abl_rkt_1}
\end{table}
\footnotetext{The number next to the cutmix word stands by the occurrence probability. \label{fn:cutmix_1}}

In the Iterations experiment, we ensured that 14k iterations were short for training, and it proved necessary to push the iterations up to 200k. In the next experiment, Single Stage Training, we discarded any benefit from the two-stage training regime or WCE. 

Later, in the Data Augmentation experiment, we adopted cropping of (224, 224), random edge resizing with scale [0.78, 2], and color augmentation with grayscale and jitter of 0.27. Cropping alone provided the best result. The addition of resizing or color augmentation worsened the results, cutmix alone made little impact, and no augmentation hypothesis had a terrible performance, revealing the need for data augmentation.

Subsequently, we tried a deeper ResNet version, which performed better. After, we tried ResNet variants: Res2Net, which uses different scales within the ResNet module; ResNeSt, which adds an attention mechanism; and ResNeXt, which works with grouped convolutions. No variant lifted the performance.

\subsection{Part 2 - Tuning DeepLabV3+}

In this section, we cover seven more experiments: Model Architecture (A), Output Stride (A), Max-Pooling Removal (A), Transposed Convolutions (A), Hybrid Local Feature Extractor (A), Cutmix (T), and Optimizer (T). See Table \ref{tab:abl_rkt_2}. Table \ref{tab:abl_rkt_2} experiment follows the best setup from Table \ref{tab:abl_rkt_1} experiments. 

In the model architecture experiment, we tried DeepLabV3+ (DL3+). It did not outperform U-Net in the first trials. However, after controlling the usage of the max-pooling (MP) layer and the output stride (OS), it reached a higher performance. In the OS experiment, in which we control the dimension ratio between the input and the encoder's outcome, there was a clear trend that reducing OS increases performance. Later, in the experiment that we suggest the MP removal, this trend reverted, and a higher OS reached the best performance. This behavior's change in OS is interpreted in the next section.

\begin{table}[hbt!]
\centering
\caption{Part 2 of PISSS.
Abbreviations: Arch (Architecture), wo/ MP (without MP layer), R50/101 (ResNet-50/101).}
\vspace{.15cm}
\begin{tabular}{l|l|l|l|l|l}
\hline
Exp. & Architecture & Encoder & OS & wo/ MP & mIoU \\
\hline
A: Arch & U-Net & R101 & - &  - & \textbf{76.5} \\
A: Arch & DL3+ & R101 & 16 &  - & 75.6 \\
A: Arch & DL3+ & R50 & 16 &  - & 75.6 \\
\hline
A: OS & DL3+ & R50 & 8 &  - & 75.7 \\
A: OS & DL3+ & R50 & 4 &  - & 76.3 \\
\hline
A: wo/ MP & DL3+ & R50 & 4 &  \checkmark & 76.2 \\
A: wo/ MP & DL3+ & R50 & 8 &  \checkmark & 76.8 \\
A: wo/ MP & DL3+ & R50 & 16 &  \checkmark & \textbf{76.9} \\
\hline
\end{tabular}
\label{tab:abl_rkt_2}
\end{table}

\subsubsection{Interpreting DeepLabV3+ Patterns} \label{dv3_patt}

\begin{figure}[hbt!]
\centering
\includegraphics[width=\textwidth]{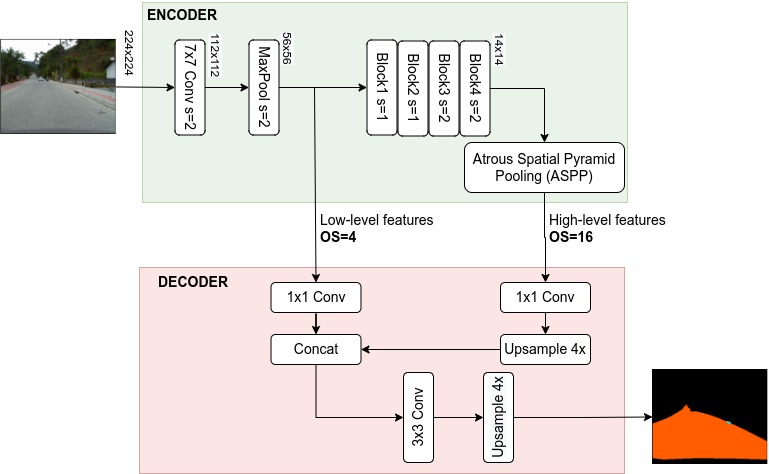}
\caption{Low and high-level features connections in DeepLabV3+.}
\label{fig:Deeplabv3pArch}
\end{figure}

It is essential to point out that the decoder of DL3+ concatenates high-level (HL) features from the ASPP block and low-level (LL) features from the ResNet's stem. See Figure \ref{fig:Deeplabv3pArch} for a vanilla example of the architecture and the control of the dimensions. The OS experiment's parameter just controls the final HL feature dimensions. On the other hand, the usage of the MP layer controls the LL dimensions; for example, without the MP layer, the LL features come just after the 7x7 conv layer with a stride of 2; otherwise, the features come just after the MP layer with a stride of 4.

Removing the MP layer positively impacted the OS of 8 and 16, whereas it did not matter for OS 4. A smaller LL stride seems more crucial than the HL stride for DL3+. Thus, when concatenating a lower LL stride with a higher HL one, it joins the best of both contexts, explaining the results of Table \ref{tab:abl_rkt_2}. 

\begin{figure}[hbt!]
\centering
\includegraphics[width=\textwidth]{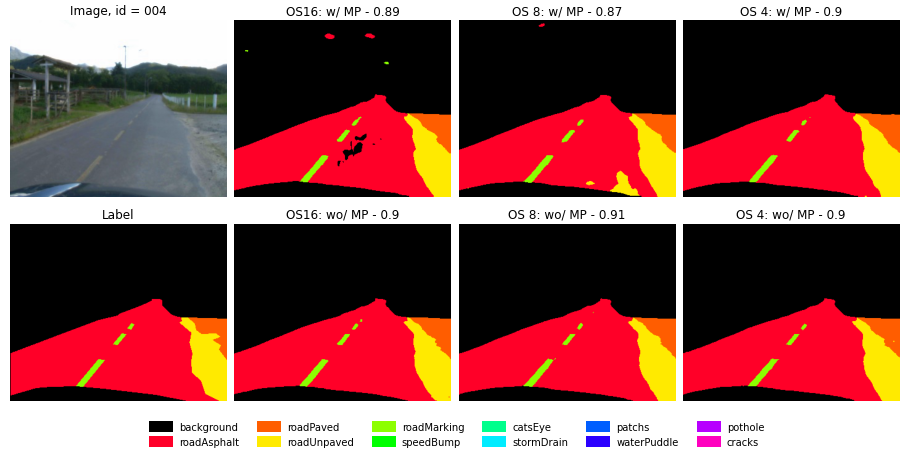}
\vspace*{-6mm}
\caption{Comparing results of distinct OS w/ or wo/ MP layer. The subtitles have the OS and MP states and the prediction IoU result. Removing the MP layer avoids early spatial information loss for extracting small object features.}
\label{fig:ch4_abl_woMP_004}
\end{figure}

\begin{figure}[hbt!]
\centering
\includegraphics[width=\textwidth]{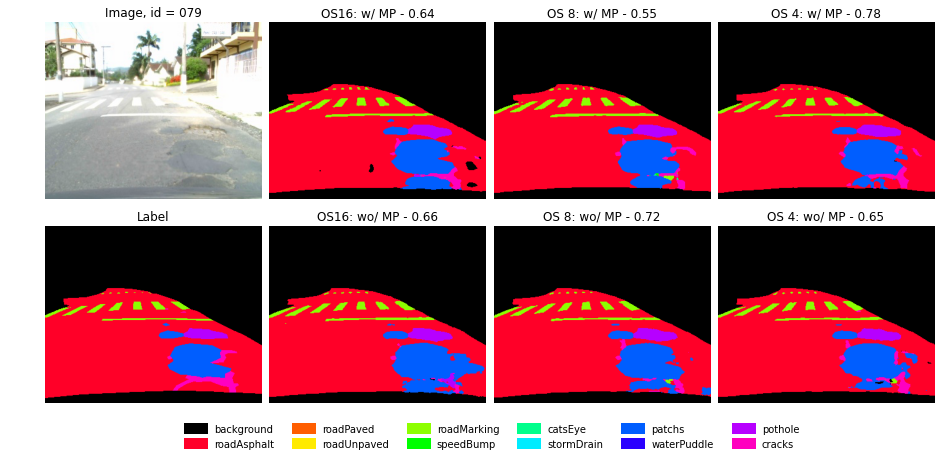}
\caption{Comparing results of distinct OS w/ or wo/ MP layer.}
\label{fig:ch4_abl_woMP_656}
\end{figure}

Another pattern found that is quantitatively unnoticeable but qualitatively impactful was that, with a higher OS, the model randomly predicts small background blobs over the road and small road blobs over the background. This problem gradually vanished until the OS decreased to 4. Besides, the problem was also solved after the MP removal; see Figures \ref{fig:ch4_abl_woMP_004} and \ref{fig:ch4_abl_woMP_656}.

\subsection{Part 3 - Fancy Approaches and Cutmix}

In this section, we cover four more experiments: Transposed Convolutions (A), Hybrid Local Feature Extractor (A), Cutmix (T), and Optimizer (T). See Table \ref{tab:abl_rkt_3}. 

We tried two fancy approaches, and none brought a performance improvement. First, we tried replacing the non-parametric upsampling with a transposed convolution layer. Next, we implemented a hybrid local feature extractor (HLFE) that joins the digressive dilation rates \cite{hamaguchi2018effective} and the hybrid dilation rates \cite{wang2018understanding}. For HLFE, we implemented the following dilation rates of [1, 3, 5, 5, 3, 1] for block 3 and [1, 3, 1] for block 4 of the ResNet. It is impossible to try HLFE for OS 16 as it has no dilation rate. 

We also tried SGD over Adam, and it neither presented any improvement. The SGD setup had LR of 1e-2 with a linear warm-up of 5k iterations, poly LR decay, and momentum of 0.9. On the other hand, trying the cutmix augmentation together with cropping had a quite effective impact, raising performance from 76.9 to 78.2 mIoU. We tested it with probability occurrence of 50\% and 80\%.

\begin{table}[hbt!]
\centering
\caption{Part 3 of PISSS.
Abbreviations: ConvT (Transposed Convolution).}
\vspace{.15cm}
\begin{tabular}{l|l|l|l|l|l|l}
\hline
Exp. & OS & ConvT & HLFE & Cutmix & SGD & mIoU \\
\hline
A: ConvT & 16 & - & - & - & - & 76.9 \\
A: ConvT & 16 & \checkmark & - & - & - & 76.0 \\
\hline
A: HLFE & 4 & - & \checkmark & - & - & 76.6 \\
A: HLFE & 8 & - & \checkmark & - & - & 76.7 \\
\hline
T: Cutmix & 16 & - & - & 50\% & - & 77.1 \\
T: Cutmix & 16 & - & - & 80\% & - & \textbf{78.2} \\
\hline
T: Optimizer & 16 & - & - & 80\% & \checkmark & 76.2 \\

\hline
\end{tabular}
\label{tab:abl_rkt_3}
\end{table}

\subsection{Part 4 - Loss Functions}

We tried the surrogate losses of mIoU and dice. It did not present any advantage on a well-calibrated training setup, although it does help the baseline simpler training setup, see Table \ref{tab:abl_rkt_4}. We noticed the surrogate losses alone degrade performance, whereas CE acts like a proxy for mIoU, optimizing it even better than its loss.

\begin{table}[hbt!]
\centering
\caption{Part 4 of PISSS. Abbreviations: R34/50 (ResNet-34/101).}
\vspace{.15cm}
\begin{tabular}{l|l|l|l|l|l|l|l}
\hline
Exp. & Arch. & Enc. & OS & wo/ MP & Aug. & Losses & mIoU \\
\hline
T: Losses & DL3+ & R50 & 16 & \checkmark & Crop+Cutmix80 & WCE &  68.9 \\
T: Losses & DL3+ & R50 & 16 & \checkmark & Crop+Cutmix80 & mIoU &  75.1 \\
T: Losses & DL3+ & R50 & 16 & \checkmark & Crop+Cutmix80 & dice &  75.6 \\
T: Losses & DL3+ & R50 & 16 & \checkmark & Crop+Cutmix80 & CE+mIoU &  77.0 \\
T: Losses & DL3+ & R50 & 16 & \checkmark & Crop+Cutmix80 & CE+dice &  77.4 \\
T: Losses & DL3+ & R50 & 16 & \checkmark & Crop+Cutmix80 & CE &  \textbf{78.2} \\
\hline
\hline
\multicolumn{8}{|l|}{\textit{Baseline}} \\
T: Losses & U-Net & R34 & - & - & GeomRTK & mIoU &  69.8 \\
T: Losses & U-Net & R34 & - & - & GeomRTK & dice &  72.0 \\
T: Losses & U-Net & R34 & - & - & GeomRTK & WCE &  73.2 \\
T: Losses & U-Net & R34 & - & - & GeomRTK & CE &  73.9 \\
T: Losses & U-Net & R34 & - & - & GeomRTK & CE+dice &  74.7 \\
T: Losses & U-Net & R34 & - & - & GeomRTK & CE+mIoU &  \textbf{74.9} \\
\hline
\end{tabular}
\label{tab:abl_rkt_4}
\end{table}

\subsection{Part 5 - Prediction Ensemble}

In contrast with the evaluation methodology applied so far, this experiment counts on the evaluation metrics of a single checkpoint (ckpt), either from the last training step or from the step with the best validation result; hence, the values presented in Table \ref{tab:abl_rkt_5} diverge from the previously reported values. We adopted the 288x224 and 448x352 resolutions for multiscale predictions besides the 352x288 native one. Ultimately, the flipped ensemble got the best results, with 78.9 and 79.8 mIoU for the last and best checkpoints.

\begin{table}[hbt!]
\centering
\caption{Part 5 of the PISSS experiments.}
\vspace{.15cm}
\begin{tabular}{l|l|l|l}
\hline
Exp. & Strategy & Last ckpt & Best ckpt \\
\hline
P: Voting Ensemble & Single Prediction & 77.4 & 79.3 \\
P: Voting Ensemble & MultiScale+Flipped & 78.4 & 79.3 \\
P: Voting Ensemble & Flipped & 78.9 & 79.8 \\
\hline
\end{tabular}
\label{tab:abl_rkt_5}
\end{table}

\section{RTK Experiments' Analysis}\label{sec_rtkanalysis}

This section first analyzes how the classes' size and pixel quantity impact performance and, next, shows what the trained model looks for in each class. The results used for the analysis are from the best PISSS hypothesis, i.e., the checkpoint from the Cutmix (T) experiment.

\subsection{Performance by Classes and Groups}

In Figure \ref{fig:ch4_relation_miou}, we see a clear correlation between the mIoU metric and the object size, confirming a mIoU bias toward big objects, also pointed out in \cite{cordts2016cityscapes}. Furthermore, the worst performances by class are from cracks, water puddles, and cats' eyes, either classes of tiny objects or undefined shapes.

\begin{center}
\begin{tabular}{cc}
\includegraphics[width=.49\textwidth]{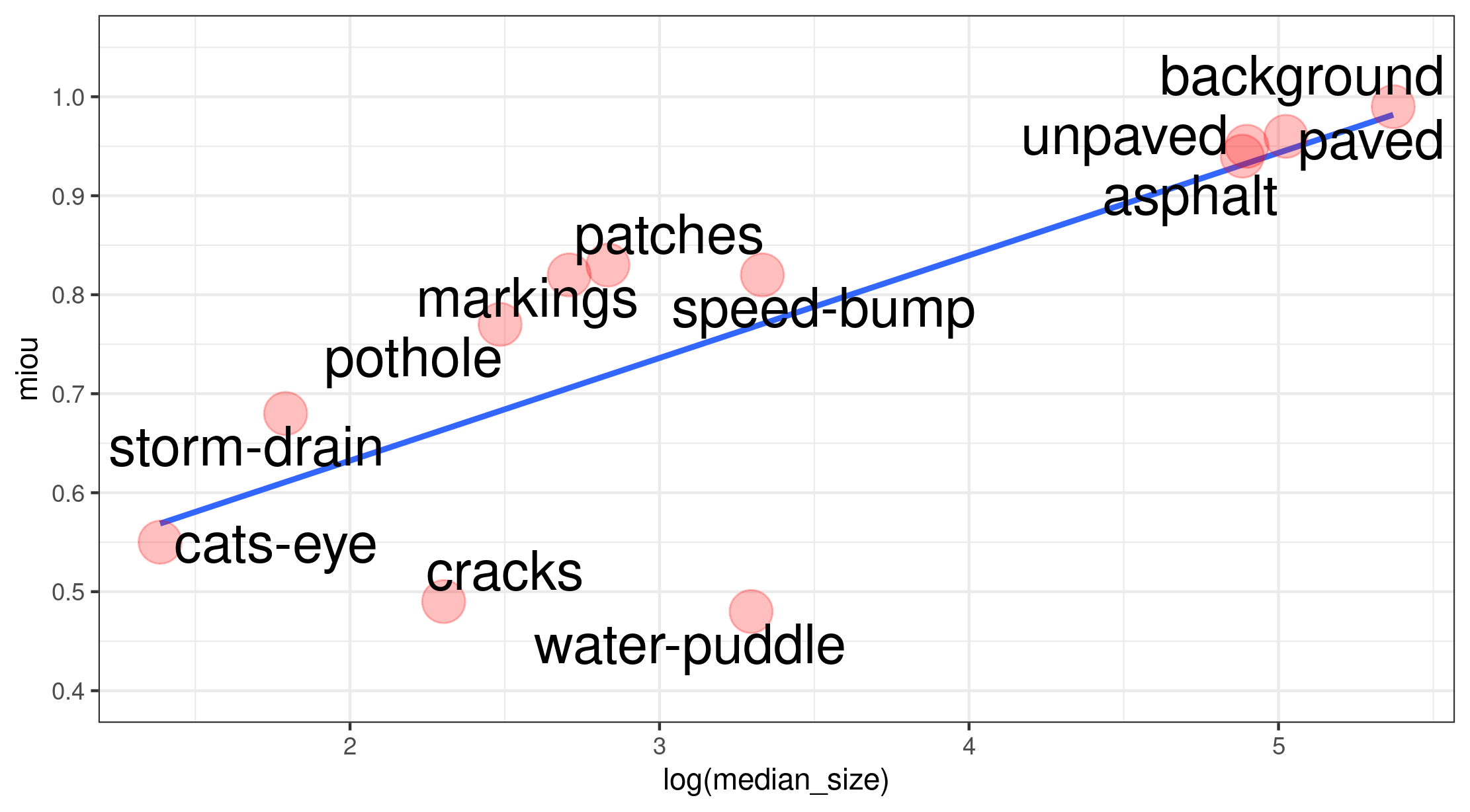}  & 
\includegraphics[width=.49\textwidth]{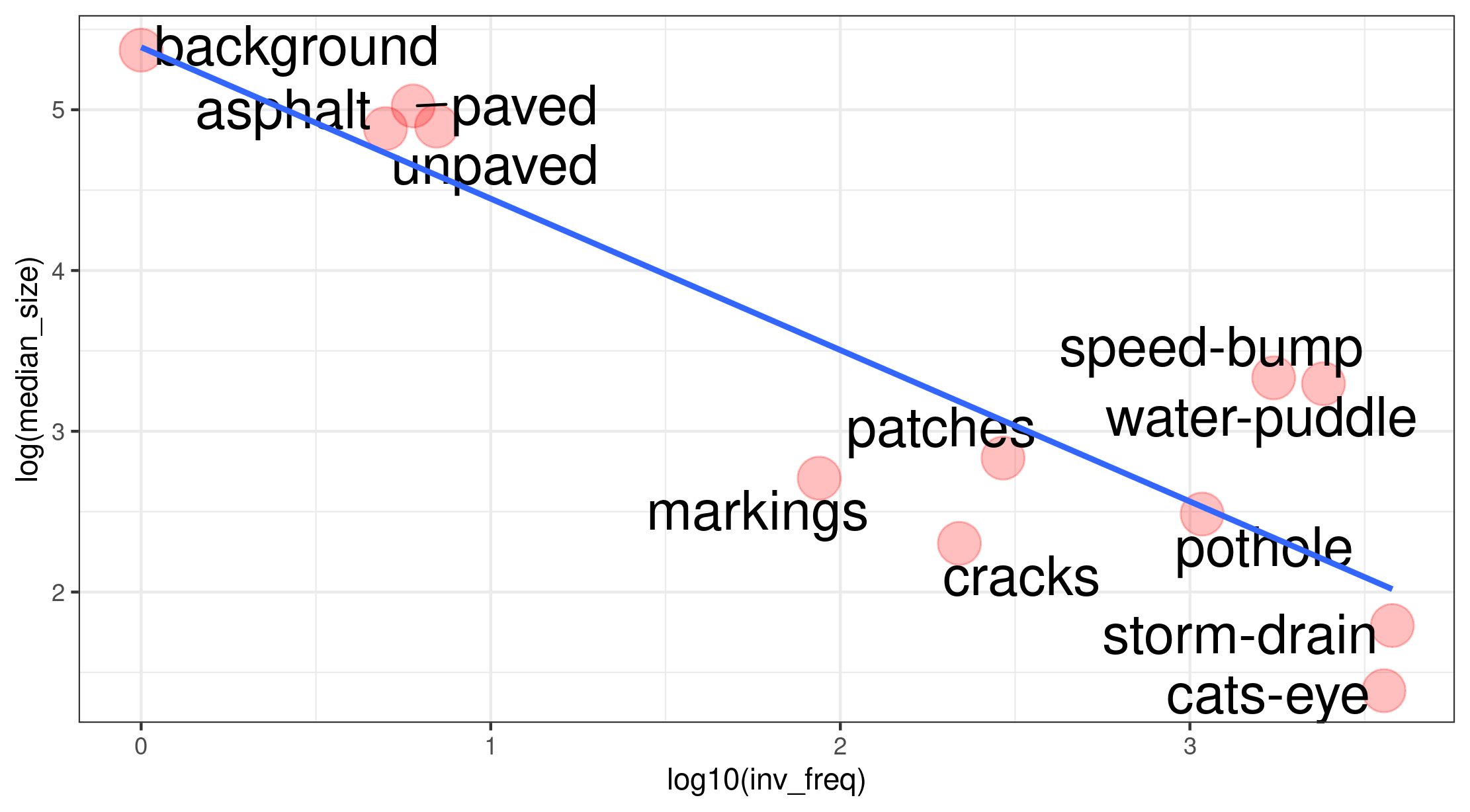} \\
 a) Impact of class edge size\footnotemark[1]. & b) Impact of inverse class frequency\footnotemark[2].\\
\end{tabular}
\captionof{figure}{Relation between class characteristics and mIoU.}
\label{fig:ch4_relation_miou}
\end{center}
\footnotetext[1]{The median size of the shortest edge of the class object.}
\footnotetext[2]{The ratio of the number of pixels between the most popular class and the class $i$.}

\subsection{What does the Neural Network look for?}

\begin{figure}[hbt!]
\centering
\includegraphics[width=\textwidth]{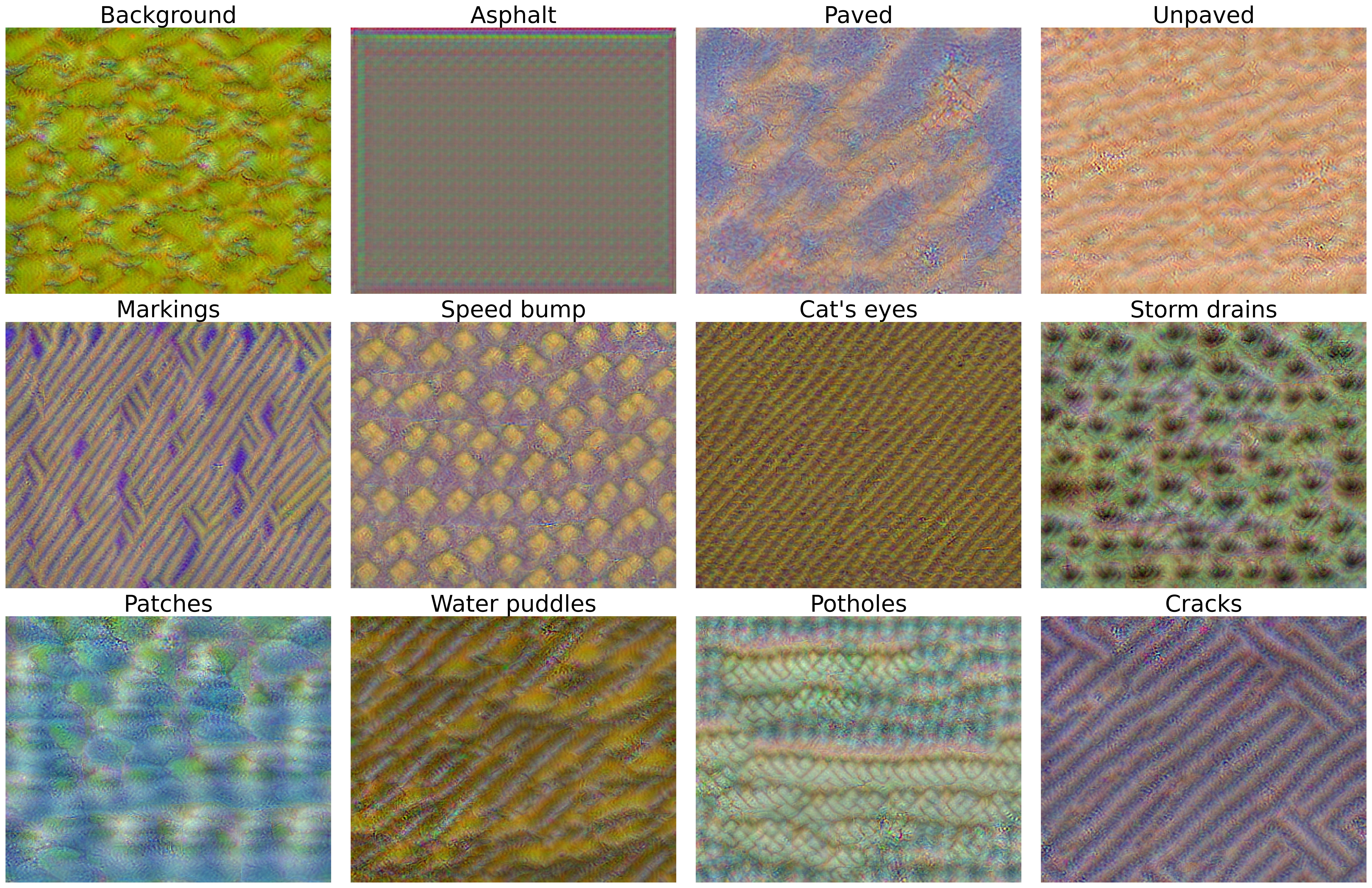}
\caption{Classes Optimized Inputs.}
\label{fig:RTK_optimuminputs}
\end{figure}

One way to understand how the neural network perceives a dataset category is by optimizing the input neurons to maximize output probability. We optimized the network's inputs using gradient ascent, shown in Figure \ref{fig:RTK_optimuminputs}. It highly noticed different textures, color distribution, and geometric patterns attached to each class; for example, storm-drain presents black holes, road-paved has the presence of polygon structures, and road-asphalt also seems to capture the cracks that happen over the road surface.

\section{Applying PISSS on TAS500}
The TAS500 dataset \cite{metzger2021fine} is a dataset from 2021 that meets unstructured environments with annotations of fine-grained vegetation and terrain classes to distinct drivable surfaces and natural obstacles. TAS500 has high-res (HR) images of 2026x620, which turned prohibitive the experiments of \textit{MP layer removal} and \textit{transposed convolutions} due to the GPU 16GB memory. Besides, the batch size had to be reduced to 4. Moreover, the images were trained with cropped parts of 1024x512, a standard practice for training HR images. 

Furthermore, we skipped the baseline experiments and only applied a subset of the PISSS hypotheses used on RTK. So the experiments' setup started with DeepLabV3+, ResNet50, OS 16, Adam with LR of 5e-
5 (a reduced value from the 1e-4 in the RTK experiments, given the reduced batch size). For validation evaluation, we followed the same methodology of averaging the results of the last ten steps. See Tables \ref{tab:abl_tas500} and \ref{tab:abl_tas500_2}.

We found that resizing as data augmentation, CE+dice loss, SGD, and multiscale ensemble prediction were fundamental for raising the TAS500 benchmark. The PISSS raised the validation set results from 65.4 to 74.7 mIoU. Furthermore, we checked our best hypothesis model on the Outdoor Semantic Segmentation Challenge, and it reached 68.8\footnote{Results with user \textit{slow} on \url{https://codalab.lisn.upsaclay.fr/competitions/5637\#results.}} mIoU, surpassing the 2021 1st place of 67.5 mIoU. 

\begin{table}[hbt!]
\centering
\caption{Summary of the PISSS experiments on TAS500. Abbreviations: Res (Resizing), CM (Cutmix).}
\vspace{.15cm}
\begin{tabular}{l|l|l|l|l|l|l|l}
\hline
Exp. & OS & Aug. & HLFE & CM & Losses & Optim & mIoU \\
\hline
T: DataAug & 16 & Crop+Color & - & - & CE & Adam & 65.4 \\
T: DataAug & 16 & Crop & - & - & CE & Adam & 67.6 \\
T: DataAug & 16 & Res+Crop+Color & - & - & CE & Adam & 69.4 \\
T: DataAug & 16 & Res+Crop & - & - & CE & Adam & \textbf{69.8} \\
\hline
A: OS & 32 & Res+Crop & - & - & CE & Adam & 65.2 \\
A: OS & 8 & Res+Crop & - & - & CE & Adam & 66.5 \\
\hline
T: HLFE & 16 & Res+Crop & \checkmark & - & CE & Adam & 69.8 \\
\hline
T: Cutmix & 16 & Res+Crop & - & 80\% & CE & Adam & 69.1 \\
\hline
T: Losses & 16 & Res+Crop & - & - & CE+dice & SGD & \textbf{70.8} \\

\hline
\end{tabular}
\label{tab:abl_tas500}
\end{table}

\begin{table}[hbt!]
\centering
\caption{Prediction Ensemble on TAS500.}
\vspace{.15cm}
\begin{tabular}{l|l|l}
\hline
Exp. & Strategy & Best ckpt \\
\hline
P: Voting Ensemble & Single Prediction & 73.1 \\
P: Voting Ensemble & Flipped & 73.6 \\
P: Voting Ensemble & MultiScale+Flipped & 74.9 \\
\hline
\end{tabular}
\label{tab:abl_tas500_2}
\end{table}

\section{Discussion and Findings} \label{sec_discussions}

Although PISSS worked for RTK and TAS500 datasets, a very different subset of hypotheses was the best in each case. For RTK, when using the MP layer, cutmix, OS 4, CE loss, and Adam worked better; while for TAS500, resizing, OS 16, CE+dice loss, and SGD worked better. The dissimilarity between these two training setups endorses the need for a custom solution, which also corroborates with the ideas \cite{wightman2021resnet} that there is no ideal training procedural for all models.

The cutmix augmentation had a meaningful gain of 1.3 mIoU for RTK, but it did not help TAS500; we suppose that cutmix is more helpful for tricky scenes of rough transitions between the road surfaces and damage classes. 

Finally, we summarize all findings on the following items:

\begin{itemize}
  \item Segmenting small objects is problematic if the features' dimensions are reduced before producing deep features; the model tends to overpredict small objects.
  \item The hardest damage classes to segment based on the RTK results are cracks, water puddles, and cat's eyes. 
  \item We confirmed the mIoU towards big objects.
  \item CE optimizes the mIoU metric better than mIoU and dice losses.
  \item An experiment performance gain depends on the initial setup that it was tested.
  \item The high dissimilarity between the best training setup for each dataset endorses the need for custom solutions.
  \item Conventional setups usually bring better results than fancy procedures and should be the first attempt.
\end{itemize}

\section{Conclusions}\label{sec_conclusions}

PISSS was effective for RTK and TAS500, reaching SOTA results and showing the importance of a well-tuning training setup besides the bare choice of a neural network architecture. Moreover, we warned of the pitfalls of early stride on convolutional networks when working with tiny objects and road damage in low-res images. We also highlighted the potential cause of problems for false-positive blobs on DeepLabV3+ predictions due to early large strides.

\begin{credits}
\subsubsection{\discintname}
The authors have no competing interests to declare that are relevant to the content of this article. 
\end{credits}
%
%
%
\bibliographystyle{splncs04}
\bibliography{sbc-template}

\begin{thebibliography}{10}
\providecommand{\url}[1]{\texttt{#1}}
\providecommand{\urlprefix}{URL }
\providecommand{\doi}[1]{https://doi.org/#1}

\bibitem{bello2021revisiting}
Bello, I., Fedus, W., Du, X., Cubuk, E.D., Srinivas, A., Lin, T.Y., Shlens, J., Zoph, B.: Revisiting resnets: Improved training and scaling strategies. Advances in Neural Information Processing Systems  \textbf{34},  22614--22627 (2021)

\bibitem{brostow2009semantic}
Brostow, G.J., Fauqueur, J., Cipolla, R.: Semantic object classes in video: A high-definition ground truth database. Pattern Recognition Letters  \textbf{30}(2),  88--97 (2009)

\bibitem{chen2017deeplab}
Chen, L.C., Papandreou, G., Kokkinos, I., Murphy, K., Yuille, A.L.: Deeplab: Semantic image segmentation with deep convolutional nets, atrous convolution, and fully connected crfs. IEEE transactions on pattern analysis and machine intelligence  \textbf{40}(4),  834--848 (2017)

\bibitem{chen2016attention}
Chen, L.C., Yang, Y., Wang, J., Xu, W., Yuille, A.L.: Attention to scale: Scale-aware semantic image segmentation. In: Proceedings of the IEEE conference on computer vision and pattern recognition. pp. 3640--3649 (2016)

\bibitem{chen2018encoder}
Chen, L.C., Zhu, Y., Papandreou, G., Schroff, F., Adam, H.: Encoder-decoder with atrous separable convolution for semantic image segmentation. In: Proceedings of the European conference on computer vision (ECCV). pp. 801--818 (2018)

\bibitem{CNT2021}
CNT: Pesquisa CNT de rodovias 2021. SEST SENAT (2021), \url{https://pesquisarodovias.cnt.org.br/downloads/ultimaversao/}

\bibitem{cordts2016cityscapes}
Cordts, M., Omran, M., Ramos, S., Rehfeld, T., Enzweiler, M., Benenson, R., Franke, U., Roth, S., Schiele, B.: The cityscapes dataset for semantic urban scene understanding. In: Proceedings of the IEEE conference on computer vision and pattern recognition. pp. 3213--3223 (2016)

\bibitem{gao2019res2net}
Gao, S.H., Cheng, M.M., Zhao, K., Zhang, X.Y., Yang, M.H., Torr, P.: Res2net: A new multi-scale backbone architecture. IEEE transactions on pattern analysis and machine intelligence  \textbf{43}(2),  652--662 (2019)

\bibitem{geiger2013vision}
Geiger, A., Lenz, P., Stiller, C., Urtasun, R.: Vision meets robotics: The kitti dataset. The International Journal of Robotics Research  \textbf{32}(11),  1231--1237 (2013)

\bibitem{hamaguchi2018effective}
Hamaguchi, R., Fujita, A., Nemoto, K., Imaizumi, T., Hikosaka, S.: Effective use of dilated convolutions for segmenting small object instances in remote sensing imagery. In: 2018 IEEE winter conference on applications of computer vision (WACV). pp. 1442--1450. IEEE (2018)

\bibitem{he2019bag}
He, T., Zhang, Z., Zhang, H., Zhang, Z., Xie, J., Li, M.: Bag of tricks for image classification with convolutional neural networks. In: Proceedings of the IEEE/CVF Conference on Computer Vision and Pattern Recognition. pp. 558--567 (2019)

\bibitem{hu2017finding}
Hu, P., Ramanan, D.: Finding tiny faces. In: Proceedings of the IEEE conference on computer vision and pattern recognition. pp. 951--959 (2017)

\bibitem{keskar2017improving}
Keskar, N.S., Socher, R.: Improving generalization performance by switching from adam to sgd. arXiv preprint arXiv:1712.07628  (2017)

\bibitem{kirillov2020pointrend}
Kirillov, A., Wu, Y., He, K., Girshick, R.: Pointrend: Image segmentation as rendering. In: Proceedings of the IEEE/CVF conference on computer vision and pattern recognition. pp. 9799--9808 (2020)

\bibitem{li2019road}
Li, Y., Peng, B., He, L., Fan, K., Li, Z., Tong, L.: Road extraction from unmanned aerial vehicle remote sensing images based on improved neural networks. Sensors  \textbf{19}(19), ~4115 (2019)

\bibitem{metzger2021fine}
Metzger, K.A., Mortimer, P., Wuensche, H.J.: A fine-grained dataset and its efficient semantic segmentation for unstructured driving scenarios. In: 2020 25th International Conference on Pattern Recognition (ICPR). pp. 7892--7899. IEEE (2021)

\bibitem{rateke2019road}
Rateke, T., Justen, K.A., Von~Wangenheim, A.: Road surface classification with images captured from low-cost camera-road traversing knowledge (rtk) dataset. Revista de Inform{\'a}tica Te{\'o}rica e Aplicada  \textbf{26}(3),  50--64 (2019)

\bibitem{rateke2021road}
Rateke, T., Von~Wangenheim, A.: Road surface detection and differentiation considering surface damages. Autonomous Robots  \textbf{45}(2),  299--312 (2021)

\bibitem{shen2017dsod}
Shen, Z., Liu, Z., Li, J., Jiang, Y.G., Chen, Y., Xue, X.: Dsod: Learning deeply supervised object detectors from scratch. In: Proceedings of the IEEE international conference on computer vision. pp. 1919--1927 (2017)

\bibitem{shinzato2016carina}
Shinzato, P.Y., dos Santos, T.C., Rosero, L.A., Ridel, D.A., Massera, C.M., Alencar, F., Batista, M.P., Hata, A.Y., Os{\'o}rio, F.S., Wolf, D.F.: Carina dataset: An emerging-country urban scenario benchmark for road detection systems. In: 2016 IEEE 19th international conference on intelligent transportation systems (ITSC). pp. 41--46. IEEE (2016)

\bibitem{tao2020hierarchical}
Tao, A., Sapra, K., Catanzaro, B.: Hierarchical multi-scale attention for semantic segmentation. arXiv preprint arXiv:2005.10821  (2020)

\bibitem{varma2019idd}
Varma, G., Subramanian, A., Namboodiri, A., Chandraker, M., Jawahar, C.: Idd: A dataset for exploring problems of autonomous navigation in unconstrained environments. In: 2019 IEEE Winter Conference on Applications of Computer Vision (WACV). pp. 1743--1751. IEEE (2019)

\bibitem{wang2018understanding}
Wang, P., Chen, P., Yuan, Y., Liu, D., Huang, Z., Hou, X., Cottrell, G.: Understanding convolution for semantic segmentation. In: 2018 IEEE winter conference on applications of computer vision (WACV). pp. 1451--1460. Ieee (2018)

\bibitem{wightman2021resnet}
Wightman, R., Touvron, H., J{\'e}gou, H.: Resnet strikes back: An improved training procedure in timm. arXiv preprint arXiv:2110.00476  (2021)

\bibitem{yu2021bisenet}
Yu, C., Gao, C., Wang, J., Yu, G., Shen, C., Sang, N.: Bisenet v2: Bilateral network with guided aggregation for real-time semantic segmentation. International Journal of Computer Vision  \textbf{129}(11),  3051--3068 (2021)

\bibitem{yuan2019segmentation}
Yuan, Y., Chen, X., Chen, X., Wang, J.: Segmentation transformer: Object-contextual representations for semantic segmentation. arXiv preprint arXiv:1909.11065  (2019)

\bibitem{zhao2017pyramid}
Zhao, H., Shi, J., Qi, X., Wang, X., Jia, J.: Pyramid scene parsing network. In: Proceedings of the IEEE conference on computer vision and pattern recognition. pp. 2881--2890 (2017)

\bibitem{zhou2020towards}
Zhou, P., Feng, J., Ma, C., Xiong, C., Hoi, S.C.H., et~al.: Towards theoretically understanding why sgd generalizes better than adam in deep learning. Advances in Neural Information Processing Systems  \textbf{33},  21285--21296 (2020)

\bibitem{zhou2018scale}
Zhou, P., Ni, B., Geng, C., Hu, J., Xu, Y.: Scale-transferrable object detection. In: proceedings of the IEEE conference on computer vision and pattern recognition. pp. 528--537 (2018)

\end{thebibliography}
%




\end{document}